\documentclass[10pt,twocolumn,letterpaper]{article}
\pdfoutput=1
\usepackage{cvpr}
\usepackage{times}
\usepackage{epsfig}
\usepackage{graphicx}
\usepackage{amsmath}
\usepackage{amssymb}
\usepackage{subfigure}
\usepackage{xspace}

\usepackage{multicol}
\usepackage{lipsum}
\usepackage{mwe}
\usepackage{appendix}
\usepackage{float}
\usepackage{mathabx}
\usepackage{listings}
\usepackage{soul}
\usepackage{caption}
\usepackage{fancyvrb}
\DefineVerbatimEnvironment{scenario}{Verbatim}{numbers=left,xleftmargin=5mm, numbersep=2mm, fontfamily=courier, \fontsize{7pt}{7pt}\selectfont}

\usepackage{booktabs}       % 

\newcommand{\scenic}{{\sc Scenic}\xspace}
\newcommand{\hide}[1]{}

\newcounter{myctr}

\newenvironment{myitemize}{\begin{list}{$\bullet$}
{\setlength{\topsep}{1mm}\setlength{\itemsep}{0.25mm}
\setlength{\parsep}{0.1mm}
\setlength{\itemindent}{0mm}\setlength{\partopsep}{0mm}
\setlength{\labelwidth}{15mm}
\setlength{\leftmargin}{4mm}}}{\end{list}}

\usepackage[pagebackref=true,breaklinks=true,letterpaper=true,colorlinks,bookmarks=false]{hyperref}

\cvprfinalcopy 

\title{A Programmatic and Semantic Approach to Explaining and Debugging\\ Neural Network Based Object Detectors}

\author{
  Edward Kim\\
  UC Berkeley       \\
  {\tt\small ek65@eecs.berkeley.edu}
  \and
  Gopinath, Divya\\
  SGT Inc, NASA AMES\\
  {\tt\small divya.gopinath@nasa.gov}
  \and
  Corina Pasareanu\\
  NASA AMES, Carnegie Mellon University\\
  {\tt\small corina.s.pasareanu@nasa.gov }
  \and
  Sanjit A. Seshia\\
  UC Berkeley\\
  {\tt\small sseshia@eecs.berkeley.edu}
}

\begin{document}
\twocolumn[{%
\renewcommand\twocolumn[1][]{#1}%
\maketitle
\begin{center}
    \centering
    \includegraphics[width=\textwidth]{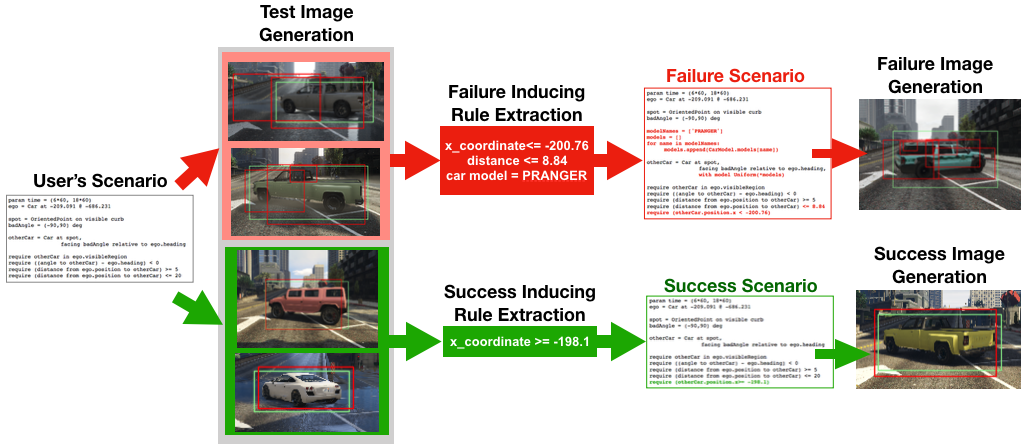}
    \captionof{figure}{Overview of the workflow proposed in this paper. The green and red bounding boxes are ground truth and prediction, respectively.}\label{fig:overview}
\end{center}%

}]

%%%%%%%%% ABSTRACT  %%%%%%%%%

\begin{abstract}
   Even as deep neural networks have become very effective for tasks in vision and perception, it remains difficult to explain and debug their behavior.
   In this paper, we present a programmatic and semantic approach to explaining, understanding, and debugging the correct and incorrect behaviors of a neural network-based perception system.
   Our approach is semantic in that it employs a high-level representation of the distribution of environment scenarios that the detector is intended to work on. It is programmatic in that scenario representation is a program in a domain-specific probabilistic programming language which can be used to generate synthetic data to test a given perception module.
   Our framework assesses the performance of a perception module to identify correct and incorrect detections, extracts rules from those results that semantically characterizes the correct and incorrect scenarios, and then specializes the probabilistic program with those rules in order to more precisely characterize the scenarios in which the perception module operates correctly or not.
   We demonstrate our results using the \scenic probabilistic programming language and a neural network-based object detector. Our experiments show that it is possible to automatically generate compact rules that significantly increase the correct detection rate (or conversely the incorrect detection rate) of the network and can thus help with understanding and debugging its behavior.
\end{abstract}

%%%%%%%%% BODY TEXT %%%%%%%%%
\pagenumbering{arabic}
\section{Introduction}
\label{sec:intro}
% \documentclass[../main.tex]{subfiles}
% \input{macros.tex}

% \begin{document}

Models produced by Machine Learning (ML) algorithms, especially {\em deep neural networks} (DNNs), have proved very effective at performing various tasks in computer vision and perception. 
Moreover, ML models are being deployed in domains where trustworthiness is a big concern, such as automotive systems~\cite{NVIDIATegra},
health care~\cite{alipanahi2015predicting}, and
cyber-security~\cite{dahl2013large}.
Research in adversarial machine learning~\cite{goodfellow-cacm18},
verification~\cite{dreossi-nfm17,seshia-arxiv16}, and testing~\cite{DBLP:conf/icse/TianPJR18} has shown that DNN-based vision/perception systems are not always robust and can be fooled, sometimes leading to unsafe situations for the overall system (e.g., autonomous vehicle).

Given this lack of robustness and potential for unsafe behavior,
it is crucial that we develop methods to better understand, debug, and characterize scenarios where DNN-based perception components fail and where they perform correctly. The emerging literature on explaining and understanding ML models provides one approach to address this concern. However, while there are several techniques proposed to explain the behavior of ML-based perception (e.g.~\cite{ActivationAtlas, NIPS2017_7062,DeepDream,Qi_2019_CVPR_Workshops,DBLP:journals/corr/SmilkovTKVW17}), almost all of them operate on the concrete input feature space of the network. For example, attribution-based methods(e.g.~\cite{DBLP:journals/corr/SundararajanTY17, DBLP:journals/corr/abs-1711-05611,SelvarajuCDVPB17}) indicate pixels in an input image that are associated with the output of a DNN on that input. These methods, while very useful, do not directly identify the higher-level ``semantic'' features of the scene that are associated with that decision; they require a human to make that judgment. Additionally, in many cases it is important to generate ``population-level'' explanations of correct/incorrect behavior on such higher-level features. For example, it would be useful to identify whether the perception module of an autonomous vehicle generally misses cars of a certain model or color, or on a particular region of a road, and leverage this knowledge to describe a high-level success/failure scenario of a perception module {\em without} the bottleneck of human intervention. \\
\indent In this paper, we present a {\em programmatic and semantic approach} to explaining and debugging 
DNN-based perception module, with a focus on object detection.
In this approach, we begin by formalizing the semantic feature space as
a distribution over a set of scenes, where a scene is a configuration of objects in the three dimensional space and semantic features are
features of the scene that capture its semantics (e.g., the position and orientation of a car, its model and color, the time of day, weather, etc.).
We then represent the semantic feature space using a program in a domain-specific programming language -- hence the term {\em programmatic}.
Given such a representation and generated data corresponding to correct and incorrect behaviors of an object detector, we seek to compute specializations of the program corresponding to those correct/incorrect behaviors. 
The specialized programs serve as interpretable representations of environment scenes that result in those correct/incorrect behaviors, enabling us to debug failure cases and to understand where the object detector succeeds.

We implement our approach using the \scenic~\cite{scenic-www,fremont-pldi19} probabilistic programming language.
Probabilistic programming has already been demonstrated to be applicable to various computer vision tasks (see, e.g.,~\cite{kulkarni2015picture}). 
\scenic is a domain-specific language used to model semantic feature spaces,
i.e., distributions over scenes. It has a generative back-end that allows one to automatically produce synthetic data when it is connected to a renderer or simulator, such as the Grand Theft Auto V (GTA-V) video game. It is
thus a particularly good fit for our approach. 
Using \scenic, we implement the workflow shown in Fig.~\ref{fig:overview}.
We begin with a \scenic program $P$ that captures a distribution that we would like our DNN-based detector to work on. Generating test data from $P$, we evaluate the performance of the detector, partitioning the test set into correct and incorrect detections.  For each partition, we use a rule extraction algorithm to generate rules over the semantic features that are highly correlated with successes/failures of the detector. Rule extraction is performed using decision tree learning and anchors~\cite{DBLP:conf/aaai/Ribeiro0G18}. 
We further propose a novel white-box approach that analyzes the neuron activation patterns of the neural network to get insights into its inner workings.  Using these activation patterns, we show how to derive semantically understandable rules over the high-level input features to characterize scenarios.
%We further propose a novel white-box approach that analyses the neuron activation patterns of the neural network to get insights into its inner workings. We show how to derive semantically understandable rules over the high-level input features to characterize scenarios that follow those patterns.

The generated rules are then used to refine $P$ yielding programs $P^+$ and $P^-$ that characterize more precisely the correct and incorrect feature spaces, respectively.
Using this framework, we evaluate DNN-based object detector for autonomous vehicles, using data generated using \scenic and GTA-V. We demonstrate that our approach is very effective, producing rules and refined programs that  significantly increase the correct detection rate (from $65.3\%$ to $89.4\%$) and incorrect detection rate (from $34.7\%$ to $87.2\%$) of the network and can thus help with understanding, debugging and retraining the network.

In summary, we make the following contributions: 
\begin{myitemize}
\item Formulation of a programming language-based semantic framework to characterize success/failure scenarios for an ML-based perception module as programs that help delineate its performance boundaries and generate new data in a principled way;
%\item The output of our framework is success and failure programs which serve as scenario descriptions delineating the performance boundaries of a perception module
\item An approach based on anchors and decision tree learning
for deriving rules for refining scenario programs;
%\item The output programs can further be used to generate more success or failure inducing data in a principled way
\item A novel white-box technique that uses activation patterns of convolutional neural networks to enhance scenario feature space refinement;
\item A data generation platform enabling research into debugging and explaining DNN-based
perception, and
\item Experimental results demonstrating that our framework is effective for a complex convolutional neural network used in autonomous driving.
\end{myitemize}

% \end{document}

\section{Background}
\label{sec:background}
% \documentclass[../main.tex]{subfiles}
% \begin{document}

\begin{table}
{\footnotesize{
\begin{tabular}{|c|p{6cm}|}
\hline
Feature & \multicolumn{1}{|c|}{Range} \\ 
\hline 
Weather & Neutral, Clear, Extrasunny, Smog, Clouds, \newline Overcast, Rain, Thunder, Clearing, Xmas, \newline
Foggy,Snowlight, Blizzard, Snow\\ 
\hline

Time  & \multicolumn{1}{|c|}{[00:00, 24:00)}\\ 
\hline

Car Model & Blista, Bus, Ninef, Asea, Baller, Bison, Buffalo, Bobcatxl, 
Dominator, Granger, Jackal, Oracle, Patriot, Pranger\\
\hline

Car Color & \multicolumn{1}{|c|}{R = [0, 255],  G = [0, 255],  B =[0, 255]} \\
\hline

Car Heading & \multicolumn{1}{|c|}{[0, 360) deg}\\
\hline

Car Position & \multicolumn{1}{|c|}{Anywhere on a road on GTA-V's map}\\
\hline
\end{tabular}
}}
\caption{Environment features and their ranges in GTA-V}\label{table:env_conditions}
\end{table}

\scenic \cite{fremont-pldi19,scenic-www} is a probabilistic programming language for scenario specification and scene generation. 
The language can be used to describe {\em environments} for autonomous systems, i.e. autonomous cars or robots. Such environments are {\em scenes}, i.e. configurations of objects and agents. \scenic allows assigning distributions to the {\em features} of the scenes, as well as hard and soft mathematical constraints over the features in the scene. Generating scenes from a \scenic program requires sampling from the distributions defined in the program. \scenic comes with efficient sampling techniques that take advantage of the structure of the \scenic program, to perform aggressive pruning.
The generated scenes are rendered into images with the help of a simulator. In this paper (and similar to \cite{fremont-pldi19}) we use the Grand Theft Auto V (GTA-V) game engine \cite{gtav} to create realistic images with a case study that uses SqueezeDet \cite{squeezeDet}, a convolutional neural network for object detection in autonomous cars. Note that the framework we put forth is not specific to SqueezeDet, and can be used with other object detectors as well. 

The semantic features that we use in our case study are described in Table \ref{table:env_conditions}. 
These features are determined and limited by the environment parameters that are controllable by users in the simulator. If distributions over these environment features are not specified in a \scenic program, then, by default, they are uniformly randomly selected from ranges shown in Table \ref{table:env_conditions}. Note that for a different application domain, we would have a different set of features.

\begin{figure}[t]
  \includegraphics[width=\columnwidth]{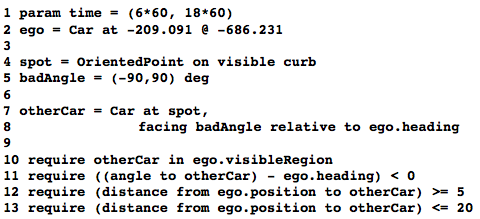}
  \caption{Example \scenic program}
%  \caption{Example \scenic program and images generated from it}
  \label{fig:scenario1}
\end{figure}

%\subsubsection{Semantics of Scenic Program}
\scenic is designed to be easily understandable, with simple and intuitive syntax. We illustrate it via an example, shown in Figure~\ref{fig:scenario1}. The formal syntax and semantics can be found in \cite{fremont-pldi19}.

As shown in Figure \ref{fig:scenario1}, the program describes a rare situation where a car is illegally intruding over a white striped traffic island to either cut in or belatedly avoid entering elevated highway. In line 1, "param time = (6*60, 18*60)" means that time of the day is uniformly randomly sampled from 6:00 to 18:00. In line 2, an ego car is placed at specific x @ y coordinate on GTA-V's map. In line 4, a spot on a traffic island that is within a visible region from a camera mounted on ego car is selected. Of all visible region of the traffic island, a spot is uniformly randomly sampled. In line 7 and 8, otherCar is placed on the spot facing -90 to 90 degree off of where ego car is facing, simulating cases when

\noindent
a car may be protruding into a traffic flow. Lastly, scenic allows users to define hard and soft constraints using "require" statements. In this scenario, all four require statements define hard constraints. In line 10, entire surface of the otherCar must be within the view region of the ego car. So, a scene where only front half of the otherCar is visible is not allowed. In line 11, the otherCar must be positioned in the right half of the ego car's visible region. In line 12 and 13, the distance of the otherCar from ego car should be 5 to 20 meters. 

% \end{document}

\section{Related Work}
\label{sec:related_work}

Most techniques that aim to provide explainability and interpretability for deep neural networks (DNNs) in the field of computer vision focus on attributing the network's decisions to portions of the input images(~\cite{NIPS2017_7062,Qi_2019_CVPR_Workshops,DBLP:journals/corr/SmilkovTKVW17,DBLP:journals/corr/SundararajanTY17, DBLP:journals/corr/abs-1711-05611}). GradCAM~\cite{SelvarajuCDVPB17} is a popular approach for interpreting CNN models  that visualizes how parts of the image affect the neural network’s output by looking into class activation maps (CAM). 
Other techniques focus on understanding the internal layers by visualizing their activation patterns ~\cite{ActivationAtlas,DeepDream}. 
Our approach, on the other hand, aims to provide characterizations at a higher level than raw image pixels, namely at the level of abstract features defined in a \scenic program. 

Rule extraction techniques either aim to represent the entire functionality of the network as a set of rules making it too complex~\cite{DeepRED} or require the presence of pre-mined set of rules~\cite{LakkarajuBL16} which would be difficult to obtain for the object detection scenario. Anchors~\cite{DBLP:conf/aaai/Ribeiro0G18}, which improves on LIME~\cite{LIME},  is closest to our work (and we discuss it in more detail later).  

Recent work aims to explain the decisions of DNNs in terms of higher-level concepts. The technique in ~\cite{KimWGCWVS18} introduces the idea of concept activation vectors,  which provide an interpretation of a neural network’s internal state in terms of
human-friendly concepts. Feature Guided Exploration~\cite{DBLP:conf/tacas/WickerHK18} aims to analyze the robustness of networks used in computer vision applications by applying perturbations over high-level input features extracted from raw images. They use object detection techniques (such as SIFT -- Scale Invariant Feature Transform) to extract the features from an image. In contrast to these techniques we directly leverage \scenic which defines the high-level features in a way that is already understandable for humans. Existing approaches typically use classification networks whose output directly corresponds to the decision being made and rely on the derivative of the output with respect to the input to calculate importance. In our application, there is no direct correlation between the output of the object detector network and the validity of the bounding boxes. Furthermore, unlike all previous work, we can use the synthesized rules to automatically generate more input instances, by refining the original \scenic program and then using it to generate data. These instances can be used to test, debug and retrain the network.

\section{Approach}
\label{sec:approach}
% \documentclass[../main.tex]{subfiles}

% \begin{document}

\begin{figure*}
  \centering
    \includegraphics[width=0.7\textwidth]{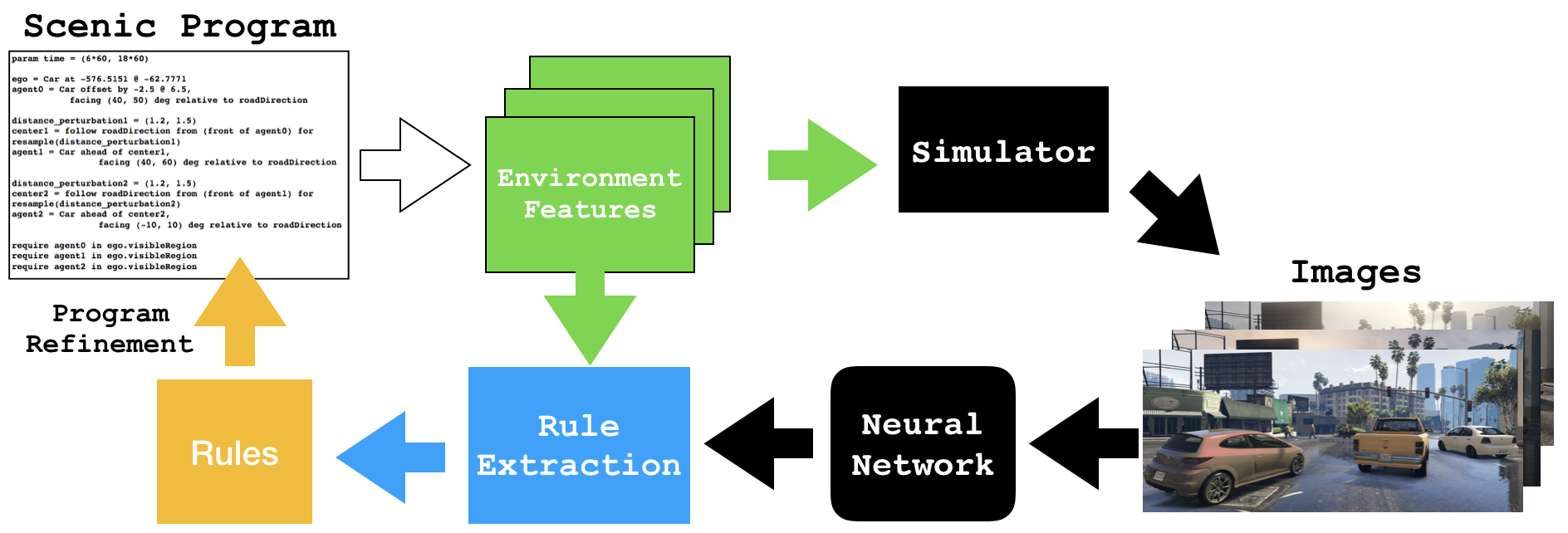}
  \caption{Analysis Pipeline}
  \label{fig:approach}
\end{figure*}

The  key  idea  of  our  approach  is  to  leverage  the  high-level semantic features formally encoded in a \scenic program to derive rules (sufficient conditions) that explain the behavior of a detection module in terms of those features. Our hypothesis is that since these features describe the important characteristics that should be present in an image and furthermore they are much fewer than the raw, low-level pixels, they should lead to small, compact rules that have a clear meaning for the developer. 

The problem that our technique aims to address can be formalized as follows. Suppose a function g defines a mapping from a feature vector, $[f_1, f_2, ..., f_n] \in D_1 \bigtimes D_2 \bigtimes ... \bigtimes D_n$, to a matrix of pixels, $m \in M$, of an image, where each $D_i$ represents the feature domain of feature $f_i$ and $M$ is a domain of $m$. Let function $h$ denote the given perception module. Finally, let $e$ be an evaluation function which compares the perception module's prediction to the ground truths, and outputs a boolean class (correct or incorrect) based on a certain performance threshold.  Given a \scenic program, according to its feature dependencies and hard and soft constraints, the feature space, $D_1 \bigtimes D_2 \bigtimes ... \bigtimes D_n$, is defined. The problem is to find the subset feature space, $d_1 \bigtimes d_2 \bigtimes ... \bigtimes d_n \subseteq D_1 \bigtimes D_2 \bigtimes ... \bigtimes D_n$ such that when we sample a certain number of features $[f_1, f_2, ..., f_n] \in d_1 \bigtimes d_2 \bigtimes ... \bigtimes d_n$, the probability that $e(h(g([f_1, f_2, ..., f_n])))$ is equal to a target class (correct or incorrect) is {\em maximized}.

A high-level overview of our analysis pipeline  is illustrated in Figure~\ref{fig:approach}. We start with a \scenic program that encodes constraints (and distributions) over high-level semantic features that are relevant for a particular application domain, in our case object detection for autonomous driving. 
Intuitively, the program (henceforth called scenario) encodes the environments that the user wants to focus on in order to test the module. Based on this scenario, \scenic generates a set of {\em feature vectors} by sampling from the specified distributions. A simulator is then used to generate a set of realistic, synthetic images (i.e. raw low-level pixel values) based on those features. 

The images are fed to the object detector. Each image is assigned a binary label, correct or incorrect, based on the performance of the object detector on the image (see Section~\ref{section:labelling}). The labels obtained for the images are mapped back to the feature vectors that led to the generation of the respective images. The result is a labeled data set that maps each high-level feature vector to the the respective label. 

We then use off-the-shelf methods to extract {\em rules} from this data set. The rule extraction is described in more detail in Sec.~\ref{sec:rule_extract}. The result is a set of rules encoding the conditions on high-level features that lead to likely correct or incorrect detection.
The obtained rules can be used to {\em refine} the \scenic program, which in turn can be sampled to generate more images that can be used to test, debug or retrain the detection module. This iterative process can continue until one obtains refined rules, and \scenic programs, of desired precision.
In the following we provide more details about our approach.

\subsection{Labelling}
\label{section:labelling}

Obtaining the label (correct/incorrect) for an image is performed  using the F1 score metric (harmonic mean of the precision and recall). This metric is commonly used in statistical analysis of binary classification.  

The F1 score is computed in the following way. For each image, the true positive (TP) is the number of ground truth bounding boxes correctly predicted by the detection module. Correctly predicted here means intersection-over-union (IoU for object detection) is greater than 0.5. The false positive (FP) is the number of predicted bounding boxes that falsely predicted ground truths. This false prediction includes duplicate predictions on one ground truth box. The false negative (FN) is the number of ground truth boxes that is not detected correctly. 
We computed the F1 score for each image, and if it is greater than a threshold, we assigned {\em correct} label; if not, {\em incorrect}. The threshold used in our experiments was 0.8.

\subsection{Rule Extraction}\label{sec:rule_extract}
\paragraph{Methods:}
We experimented with two methods, decision tree (DT) learning for classification \cite{quinlan1986induction} and anchors \cite{DBLP:conf/aaai/Ribeiro0G18},  to extract rules capturing the subspace of the feature space defined in the given \scenic program. 

Decision tree learning is commonly used to extract rules explaining the {\em global} behavior of a complex system while the anchors method is a state-of-the art technique for extracting explanation rules that are {\em locally} faithful.

Decision trees encode decisions (and their consequences) in a tree-like structure. They are highly  interpretable, provided that the trees are short. One can easily extract rules for explaining different classes, by simply following the paths through the trees and conjuncting the decisions encoded in the tree nodes.  We used the {\tt rpart} \cite{rpart} package in R software, which implements corresponding algorithm in \cite{cart84}, with default parameters.

The anchor method is a state-of-the art technique that aims to explain the behavior of complex ML models with high-precision rules called anchors, representing local, sufficient conditions for predictions. The system can efficiently compute these explanations for any black-box model with high-probability guarantees. We used the code from \url{https://github.com/marcotcr/anchor} with the default parameters. Applying the method to the object detector directly would result in anchors describing conditions on low-level pixel values, which would be difficult to interpret and use.  Instead what we want is to extract anchors in terms of high-level features. While one can use the simulator together with the object detector as the black-box model, this would be very inefficient.
Instead we built a surrogate model mapping high-level \scenic features to output labels; we used a random forest learning algorithm for this purpose as in the code. This surrogate model was then passed to the anchor method to extract the rules.

\paragraph{Blackbox vs Whitebox Analysis:} 
\label{section_4_4_2}

So far we explained how we can obtain rules when treating the detection module as a black box. We also investigated a white-box analysis, to determine whether we can exploit the information about the internal workings of the module to improve the rule inference.
The white-box analysis is one of our novel contributions in this paper. We leverage recent work  \cite{DBLP:journals/corr/abs-1904-13215} which aims to infer likely properties of neural networks. The properties are in terms of on/off activation patterns (at different internal layers) that lead to the same predictions. These patterns are computed by applying decision-tree learning over the activations observed during the execution of the network on the training or testing set.

We analyzed the architecture of the SqueezeDet network and we determined that there are three maxpool layers  which provide a natural decomposition of the network. Furthermore they have relatively low dimensionality making them a good target for property inference.

We consider activation patterns over maxpool neurons based on whether the neuron output is greater or equal to zero. A decision tree can then be learned over these patterns to fit the prediction labels. For our experiments we selected patterns from the maxpool layer 5, which turned out to be highly correlated to images that lead to correct/incorrect predictions.

Then, we augmented the assigned correct and incorrect labels with corresponding decision pattern in the following way. For example, using a decision pattern for correct labels (i.e. the decision pattern that most correlated to images with correct label), we created two sub-classes for correct class. By feeding in only images with correct label to the perception module, the images satisfying the decision pattern is re-labelled as "correct-decision-pattern," otherwise, "correct-unlabelled." Likewise, the incorrect class is augmented using a decision pattern that is most correlated to images with incorrect label. It is our intuition that the decision pattern captures more focused properties (or rules) among images belonging to a target class. Hence, we hypothesize that this label augmentation would help anchor and decision tree methods to better identify rules. 

\paragraph{Rule Selection Criteria:}
\label{section:rule_selection}
Once we extracted rules with either DT or anchors, we selected the best rule using following criteria. To best achieve our objective, first, we chose the rule with highest precision on a held-out testset of feature vectors. If there are more than one rule with equal high precision, then we chose the rule with the highest coverage (i.e. the number of feature vectors satisfying the rules). Finally, if there is still more than one rule left, then we broke the tie by choosing the most compact rule which has the least number of features. The last two criteria are established to select the most general high-precision rule.

\section{Experiments}
\label{sec:experiments}
\begin{table}[t]
{\small
\begin{tabular}{|c|p{4.5cm}|}  
\hline
Scenario \#   & \multicolumn{1}{|c|}{Rules} \\
(Baseline$\rightarrow$Rule Precision) & \\
\hline \hline
Scenario 1 & \textbf{x coordinate} $\geq$ -198.1 \\
($65.3\% \rightarrow 89.4\%$) & \\
\hline
 & \textbf{hour} $\geq$ 7.5 $\wedge$ \\
 & \textbf{weather} = all except neutral $\wedge$ \\
Scenario 2 & \textbf{car0 distance from ego} $\geq$ 11.3m $\wedge$ \\
($72.3\% \rightarrow 82.3\%$) & \textbf{car0 model} = \{Asea, Bison, Blista, 
Buffalo, Dominator, Jackal, Ninef, Oracle\}\\
\hline
Scenario 3 & \textbf{car0 red color} $\geq$ 74.5 $\wedge$ \\ 
($61.7\% \rightarrow 79.4\%$) & \textbf{car0 heading} $\geq$ 220.3 deg \\
\hline
 & \textbf{car0 model} = \{Asea, Baller, Blista, \\
Scenario 4 & Buffal, Dominator, Jackal, Ninef,\\ 
($89.6\% \rightarrow 96.2\%$) & Oracle\} \\
\hline
\end{tabular}
\\
\caption{Rules for correct behaviors of the detection module with the highest precision from Table~\ref{fig:correct_rule_precision}}
\label{table:correct_rules}
}
\end{table}

\begin{table}
{\small
\begin{tabular}{|c|p{4.5cm}|}    
\hline
Scenario \#    & \multicolumn{1}{|c|}{Rules}\\
(Baseline$\rightarrow$Rule Precision) & \\
\hline \hline
 & \textbf{x coordinate} $\leq$ -200.76 $\wedge$ \\
Scenario 1 & \textbf{distance} $\leq$ 8.84 $\wedge$ \\
($34.7\% \rightarrow 87.2\%$) & \textbf{car model} = PRANGER \\
\hline

 & \textbf{hour} $\geq$ 7.5 $\wedge$ \\
Scenario 2 & \textbf{weather} = all except Neutral $\wedge$ \\
($27.7\% \rightarrow 44.9\%$) & \textbf{car0 distance from ego} $<$ 11.3\\
\hline

 & \textbf{weather} = neutral $\wedge$\\
Scenario 3 & \textbf{agent0 heading =} $\leq$ 218.08 deg $\wedge$ \\
($38.3\% \rightarrow 83.4\%$) & \textbf{hour} $\leq$ 8.00 $\wedge$ \\
 & \textbf{car2 red color} $\leq$ 95.00 \\
\hline

 & \textbf{car0 model} = PATRIOT $\wedge$ \\
 & \textbf{car1 model} = NINEF $\wedge$ \\
Scenario 4 & \textbf{car2 model} = BALLER $\wedge$ \\
($10.4\% \rightarrow 57.3\%$) & 92.25 $<$ \textbf{car0 green color} $\leq$ 158 $\wedge$\newline
\textbf{car0 blue color} $\leq$ 84.25 $\wedge$\newline
178.00 $<$ \textbf{car2 red color} $\leq$ 224\\ 
\hline
\end{tabular}
\\
\caption{Rules for incorrect behaviors of detection module with the highest precision from Table~\ref{fig:incorrect_rule_evaluation}}
\label{table:incorrect_rules}
}
\end{table}

In this section we report on our experiments with the proposed approach on the object detector. We investigate whether we can synthesize rules that are effective in generating test inputs that increase the probability of correct/incorrect detection, thus explaining the correct/incorrect behavior of the analyzed module. We evaluate the proposed techniques along the following dimensions: decision tree (DT) vs anchor, black-box (BB) vs white-box (WB).

\subsection{Scenarios}
We experimented with our approach on four different scenarios. Scenario 1 (Figure \ref{fig:scenario1}) describes the situation where a car is illegally intruding over a white striped traffic island at the entrance of an elevated highway. 
Scenario 2 describes two-car scenario where one car occludes the ego car's view of another car at a T-junction intersection on an elevated road. 
describes scenes where other cars are merging into ego car's lane. The location in this scenario is carefully chosen such that the sun rises in front of ego car, causing a glare. 
describes a set of scenes when nearest car is abruptly switching into ego car's lane while another car on the opposite traffic direction lane is slightly intruding over the middle yellow line into ego car's lane. 

Scenario 2 describes two-car scenario where a car occludes the ego car’s view of another car at a T-junction intersection . In the \scenic program, to cause occlusion in scenes, we place an agent0 car within a visible region from ego car. Then, we place agent1 car within a close vicinity, defined by small horizontal (i.e. leftRight) and vertical (i.e. frontback) perturbations in the program, to agent0 car. The metric of these perturbations are in meters. The images from Scenario 2 are shown in Figure \ref{fig:supp_scenario2_images}. 

Scenario 3 describes scenes where three cars are merging into the ego car’s lane. The location for Scenario 3 is carefully chosen such that the sun rises in front of the ego car, causing a glare. The \scenic program describes three cars merging in a platoon-like manner where one car is following another car in front with some variations in distance between front and rear cars. The metric for distance perturbation is in meters. The images from Scenario 3 are shown in Figure \ref{fig:supp_scenario3_images}. 

Finally, Scenario 4 describes a set of scenes when the nearest car is abruptly switching into ego car’s lane while another car on the opposite traffic direction lane is slightly intruding over the middle yellow line into the ego car’s lane. Failure to detect these two cars out of four cars may potentially be safety-critical. The images from Scenario 4 are shown in Figure \ref{fig:supp_scenario4_images}. The locations of all four cars, in Scenario 4 \scenic program, are hard-coded with respect to ego car's location. The \scenic program would have become much more interpretable had we described car locations with respect to lanes. The reason we had to code in this undesirable manner is due to the simulator as illustrated later in this section.

\section{Success and Failure Scenario Descriptions} \label{sec:success_failure}
The refined \scenic programs characterizing success/failure scenarios are shown in Figure \ref{fig:supp_refined_scenario1}, \ref{fig:supp_refined_scenario2}, \ref{fig:supp_refined_scenario3}, and \ref{fig:supp_refined_scenario4}. The red/green parts of programs represent the rules automatically generated by our technique, which are cut and pasted to original \scenic programs. These success/failure inducing rules are shown in Table \ref{table:correct_rules} and \ref{table:incorrect_rules}. As aforementioned, we generated new images using \scenic programs that characterize failure scenarios. Examples of these images from failure scenarios are shown in Figure \ref{fig:supp_scenario1_failure_images}, \ref{fig:supp_scenario2_failure_images}, \ref{fig:supp_scenario3_failure_images}, and \ref{fig:supp_scenario4_failure_images}.

\subsection{Setup}
The object detector was trained on a separate set of 10,000 GTA images with one to four cars in various locations of the map producing different background scenes. 
The GTA-V simulator provided images, ground truth boxes, and values of the environment features. 

For each scenario, we generated 950 new images as a train set and another 950 new images as a test set. 
We denote the labels corresponding to the maxpool layer 5 decision pattern as p5c(correct) and p5\_ic(incorrect) and the remaining as correct\_unlabelled and incorrect\_unlabelled, respectively. We augmented the feature vector with some extra features that are not part of the feature values provided by the simulator but could help with extracting meaningful rules. For example, in Scenario 1, the distance from ego to otherCar is not part of the feature values provided by GTA-V. However, it can be computed with Euclidean distance metric using (x,y) location coordinates of ego and otherCar. Also, the difference in heading angle between ego and otherCar is also added as extra feature to represent ``badAngle'' variable in the program. 

From the train set, we extracted rules to predict each label based on the feature vectors. These rules were evaluated on the test set based on precision, recall, and F1 score metrics.  
For DT learning we adjusted the label weight to account for the uneven ratio among labels for both black-box and white-box labels. 
For the Anchors method,  we applied it on each instance of the training set until we had covered a maximum of 50 instances for every label ({correct, incorrect} for Black Box, and {p5c, p5\_ic, correct\_unlabelled, incorrect\_unlabelled} for White Box). The best anchor rule for every label is selected based on the rule selection criteria mentioned in section~\ref{section:rule_selection}.

\begin{figure}
{\small
  \centering
    \includegraphics[width=\columnwidth]{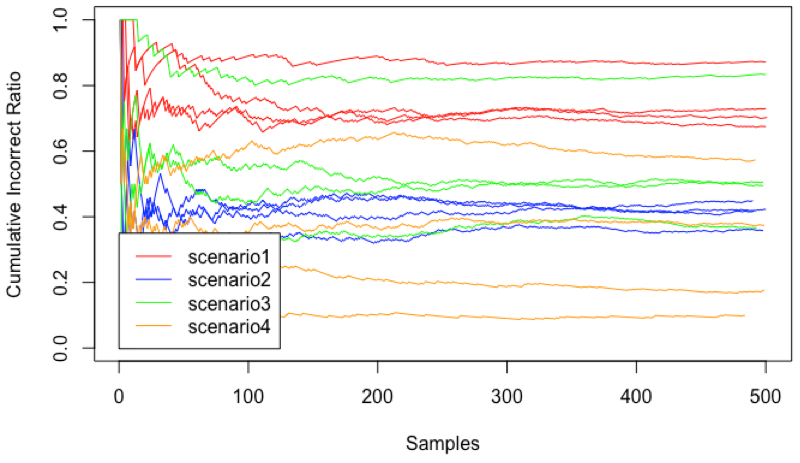}
  \caption{The cumulative ratio of incorrectly detected images generated from refined \scenic programs (using incorrect rules) stabilizes over 500 samples. Each color has four graphs representing four different rule extraction methods}
  \label{fig:incorrect_ratio_graph}
  }
\end{figure}

\begin{table}
{
\centering
\begin{tabular}{c c c c c}    \toprule
Scenario \#    & 1  & 2  & 3  & 4  \\\bottomrule \hline \hline
Correct DP & 0.626 & 0.651 & 0.514 & 0.824\\ 
Incorrect DP & 0.276 & 0.175 & 0.234 & 0.212\\
\bottomrule \hline
\end{tabular}
\\
\caption{Support for correct and incorrect decision patterns}
\label{fig:DP_support}
}
\end{table}

\begin{table}
{\small
\centering
\begin{tabular}{c c c c c}     \toprule
Scenario \#    & 1  & 2  & 3  & 4  \\\bottomrule \hline \hline
BB Decision Tree & 0.723 & 0.342 & 0.631 & 0.622\\ 
WB Decision Tree & 0.727 & 0.696 & 0.601 & 0.778\\
BB Anchor & 0.361 & 0.457 & 0.302 & 0.438\\
WB Anchor & 0.520 & 0.188 & 0.149 & 0.438\\
\bottomrule \hline
\end{tabular}
\\
\caption{F1 score of correct rules on testset}
\label{fig:f1_score_on_correct_rules}
}
\end{table}

\begin{table}
{\small
\centering
\begin{tabular}{c c c c c}     \toprule
Scenario \#    & 1  & 2  & 3  & 4  \\\bottomrule \hline \hline
Original Program & 0.653 & 0.723 & 0.617 & 0.896\\ \bottomrule \hline 
BB Decision Tree & 0.843 & 0.778 & 0.787 & 0.950\\ 
WB Decision Tree & 0.826 & 0.823 & 0.788 & 0.962\\
BB Anchor & 0.727 & 0.811 & 0.652 & 0.928\\
WB Anchor & 0.894 & 0.817 & 0.794 & 0.928\\
\bottomrule \hline
\end{tabular}
\\
\caption{Precision of correct rules on testset}
\label{fig:correct_rule_precision}
}
\end{table}

\subsection{Results}
Tables~\ref{table:correct_rules} and~\ref{table:incorrect_rules} show the best rules (wrt. precision) extracted with our proposed framework, along with the baseline correct/incorrect detection rate for each given scenario and the detection rate for the generated rules. The results indicate that indeed our framework can generate rules that increase significantly the correct and incorrect detection rate of the module. Furthermore, the generated rules are compact and easily interpretable.

For example, the rule for correct behavior for Scenario 1 is "x coordinate $\geq -198.1$." In GTA-V, at ego car's specific location, the condition on x coordinate was equivalent to the otherCar's distance from ego being greater than 11m. On the other hand, the rule for incorrect behavior for Scenario 1 requires the otherCar to be within 8.84m and its car model to be PRANGER. These rules, counter-intuitively, indicate that the object detector fails when the otherCar is close by, and performs well when located further away. 

\begin{table}
{\small
\centering
\begin{tabular}{c c c c c}     \toprule
Scenario \#    & 1  & 2  & 3  & 4  \\\bottomrule \hline \hline
Original Program & 0.347 & 0.277 & 0.383 & 0.104\\ \bottomrule \hline 
BB Decision Tree  & 0.703 & 0.418 & 0.506 & 0.375\\ 
WB Decision Tree  & 0.73 & 0.449 & 0.494 & 0.099\\
BB Anchor & 0.872 & 0.357 & 0.834 & 0.573\\
WB Anchor & 0.674 & 0.422 & 0.365 & 0.176\\
\bottomrule \hline
\end{tabular}
\\
\caption{Incorrectly detected image ratio  among five hundred new data generated from each refined \scenic program}
\label{fig:incorrect_rule_evaluation}
}
\end{table}

\begin{table}
{\small
\centering
\begin{tabular}{c c c c c} 
\hline
Scenario \#    & 1  & 2  & 3  & 4  \\\bottomrule \hline 
Feature Space Coverage  & 0.692 & 0.956 & 0.898 & 0.871 \\
\bottomrule \hline
\end{tabular}
\\
\caption{The proportion of original feature space covered by the best incorrect rule for each scenario from Table~\ref{fig:incorrect_rule_evaluation}}
\label{table:refined_program_feature_space_coverage}
}
\end{table}

\paragraph{Results for Correct Behavior:}

Tables~\ref{fig:f1_score_on_correct_rules} and~\ref{fig:correct_rule_precision} summarize the results for the rules explaining correct behavior. 
The results indicate that there are clear signals in the heavily abstracted feature space and they can be used effectively for scenario characterization via the generated high-precision rules. 

The results also indicate that DT learning extracts rules with better F1 scores for all scenarios as compared to anchors. This could be attributed to the difference in the nature of the techniques. The anchor approach aims to construct rules that have high precision in the locality of a given instance. 
Decision-trees on the other hand aim to construct global rules that discriminate one label from another. Given that a large proportion of instances were detected correctly by the analyzed module, the decision tree was able to build rules with high precision and coverage for correct behavior.

The results also highlight the benefit of using white-box information to extract rules for correct behavior. 

Table~\ref{fig:DP_support} shows the support for the decision pattern is significant (greater than 65\% on average for all scenarios). The support is defined as a correlation of the decision pattern to a specific label. Using this information to augment the labels of the dataset helped to improve the precision and F1 score of the rules (w.r.t. \scenic features) for both DT learning and anchor method.

\paragraph{Results for Incorrect Behavior:}

Tables~\ref{fig:incorrect_rule_evaluation} summarize the results for the rules explaining incorrect behavior. 
Rule derivation for incorrect behavior is more challenging than for correct behavior due to the low percentage of inputs that lead to the incorrect detection for a well trained network.

In fact the F1 scores (computed on the test set) for rules predicting incorrect behavior were too low due to very low (in some cases 0) recall values. 

To properly validate the efficacy of the generated rules, we refined the \scenic programs by encoding the rules as constraints and we generated 500 new images. We then evaluated our module's performance on these new datasets. Figure \ref{fig:incorrect_ratio_graph} justifies our choice of 500 as the number of new images that we generate for evaluation. 

All four methods contributed to more precisely identifying the subset features spaces in which the module performs worse. Specifically, Table \ref{fig:incorrect_rule_evaluation} illustrates that the black-box anchor method enhanced the generation rate of incorrectly detected images by 48\% on average in Scenarios 1, 3, and 4 compared to the baseline. This is a significant increase in the ratio of incorrectly labelled images generated from the program, providing evidence that the refined programs are more precisely characterizing the failure scenarios. 

We also note that the anchor method outperforms DT learning. This is expected,  because the anchor method extracts rules that are highly precise within a local feature space. The exception is Scenario 2. We conjecture that the reason that the anchor method did not perform better than DT learning is due to uncontrollable non-determinism in GTA-V, which generated pedestrians in close vicinity to the camera of ego car even though its \scenic program did not have any pedestrian. GTA-V non-deterministically instantiated these pedestrians, and the perception module often incorrectly predicted the pedestrians as cars. This is an issue with the GTA-V which originally was not built for data generation purpose. GTA-V does not allow users to control or eliminate these pedestrians and it does not provide features related to pedestrians during data collection process. In future work, we plan to incorporate simulators that allows a deterministic control (such as CARLA \cite{carla}) for further experimentation.

Unlike the results for correct behavior, the whitebox approach tends to perform worse than blackbox when focusing on incorrect behavior. This outcome can be attributed to very low support for decision patterns computed for incorrect behavior, with maximum of 27.6\% among the four scenarios as shown in Table~\ref{fig:DP_support}. 

However, we do observe that the white-box approach for both DT learning and anchors does, in general, enhance the ratio of incorrectly detected images as shown in Table \ref{fig:incorrect_rule_evaluation}, compared to those of the original programs.

\paragraph{Limitations:}
Our technique relies on abstracting an image with a high resolution (for instance 1920 x 1200 in our example) to a vector of a small set of semantic features. In our experiments we were able to derive compact rules with high precision and coverage. However, we do note that in other application domains, other than autonomous driving, the abstraction may lead to under-determined representation, which may not yield any noticeable patterns. Therefore, appropriate selection of a subset of essential features for a given application domain (facilitated by an appropriate definition using \scenic), is essential.  
We also note that all the \scenic programs we experimented with contained only uniform distributions. Also, for each of the scenario programs that we analyzed, we fixed the location and heading angle of the camera. In these restricted settings, we were able to extract rules that distinguished correctly detected scenes from the incorrect ones. 

% \end{document}

\section{Conclusion and Future Work}
\label{sec:conclusion}

We presented a semantic and programmatic framework for characterizing success and failure scenarios of a given perception module in the form of programs. The technique leverages the \scenic language to derive rules in terms of high-level, meaningful features and generates new inputs that conform with these rules. For future work, we plan on applying this approach to other domains, by looking into more general input distributions and transformations. %We plan to release 

\section{Acknowledgment}
\label{sec:acknowledgement}
We thank Daniel Fremont for his help on our use of \scenic, Jinkyu Kim and Taesung Park for their thorough comments, and Xiangyu Yue for interfacing GTA-V with \scenic. This project is supported by an NSF graduate fellowship (Grant\#: DGE1752814), NSF grants CNS-1545126 (VeHICaL), CNS-1739816,
and CCF-1837132, by the DARPA Assured Autonomy program, by Berkeley Deep Drive, and by Toyota under the iCyPhy center.  
This work was partially done under the NASA Ames Internship Program, 2019.

%%%%%%%%% Reference  %%%%%%%%%
{\small
\bibliographystyle{plain}
\bibliography{biblio}
}

%%%%%%%%% Appendix  %%%%%%%%%
\begin{appendices}

\begin{figure*}
  \centering
  \subfigure[Scenario 1 \scenic program]{\includegraphics[scale=0.5]{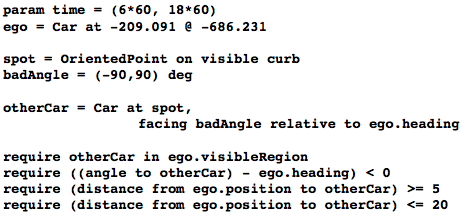}}\quad
  \subfigure[Scenario 2 \scenic program]{\includegraphics[scale=0.5]{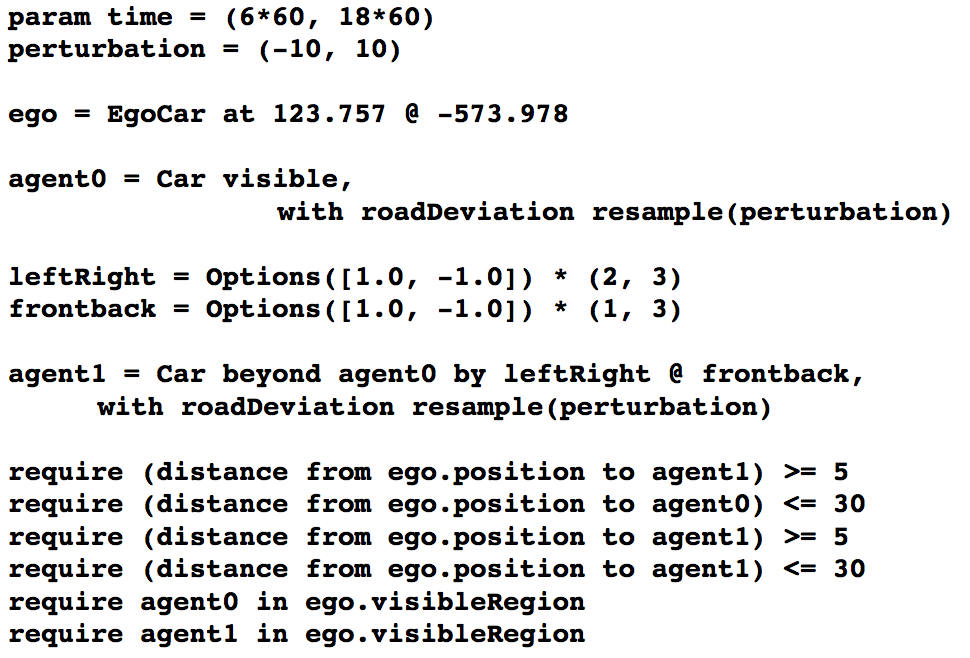}}
  \caption{\label{fig:supp_scenario1_2}%
  Scenario 1 and Scenario 2 }
\end{figure*}

\begin{figure*}
  \centering
  \subfigure[Scenario 3 \scenic program]{\includegraphics[scale=0.45]{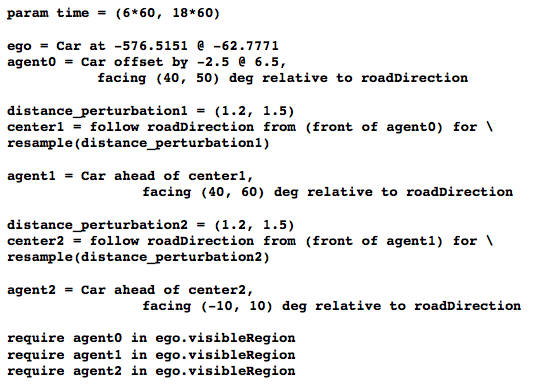}}\quad
  \subfigure[Scenario 4 \scenic program]{\includegraphics[scale=0.45]{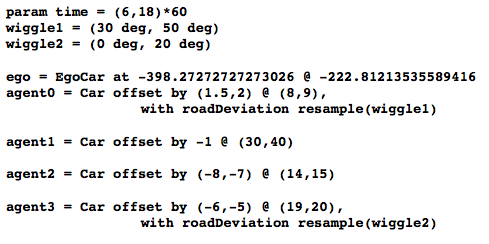}}
  
  \caption{\label{fig:supp_scenario3_4}%
  Scenario 3 and Scenario 4 }
\end{figure*}

\begin{figure*}
  \centering
    \includegraphics[width=\textwidth]{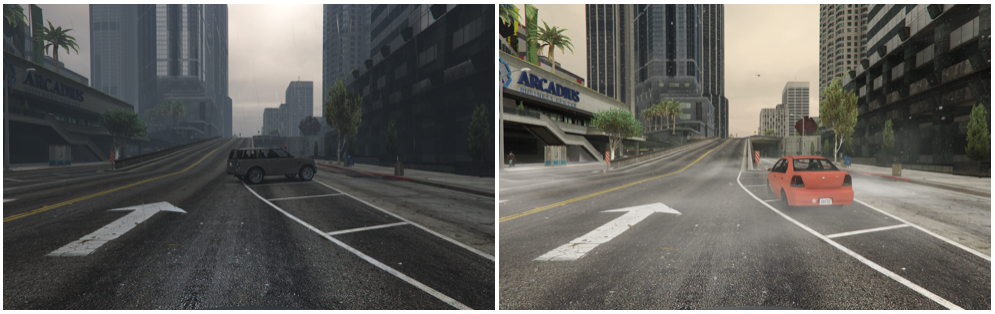}
  \caption{Images generated using Scenario 1}
  \label{fig:supp_scenario1_images}
\end{figure*}

\begin{figure*}
  \centering
    \includegraphics[width=\textwidth]{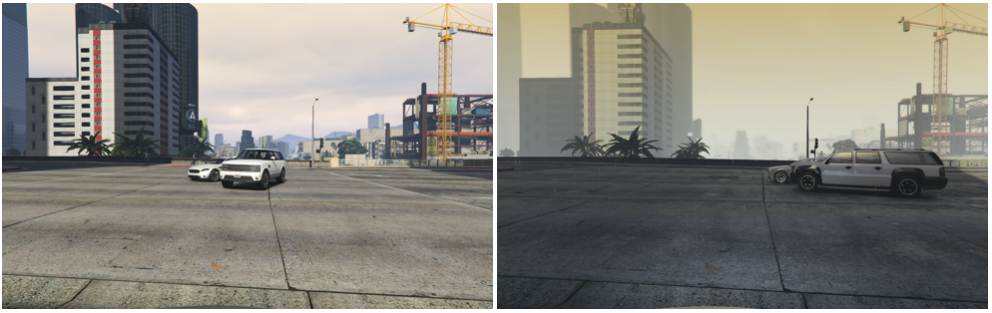}
  \caption{Images generated using Scenario 2}
  \label{fig:supp_scenario2_images}
\end{figure*}

\begin{figure*}
  \centering
    \includegraphics[width=\textwidth]{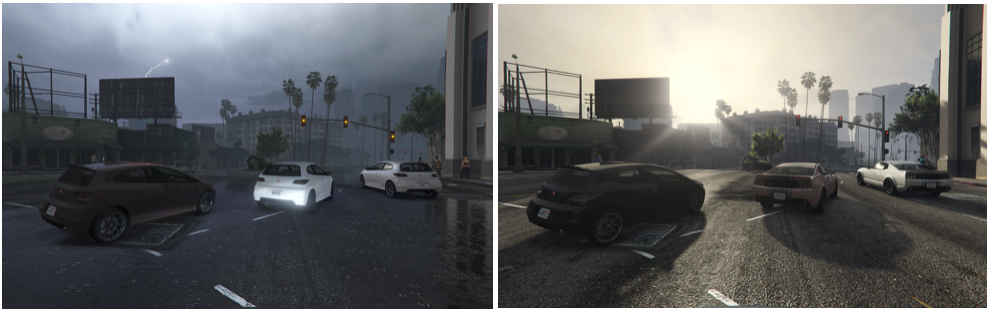}
  \caption{Images generated using Scenario 3}
  \label{fig:supp_scenario3_images}
\end{figure*}

\begin{figure*}
  \centering
    \includegraphics[width=\textwidth]{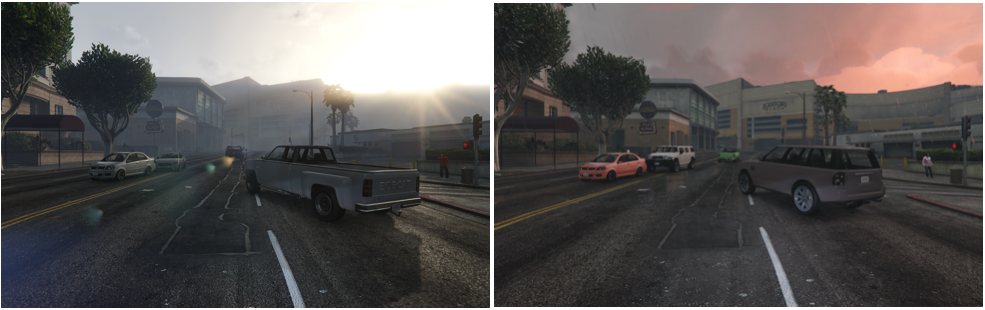}
  \caption{Images generated using Scenario 4}
  \label{fig:supp_scenario4_images}
\end{figure*}

\begin{figure*}
  \centering
  \subfigure[Refined Success Scenario 1 \scenic program]{\includegraphics[scale=0.5]{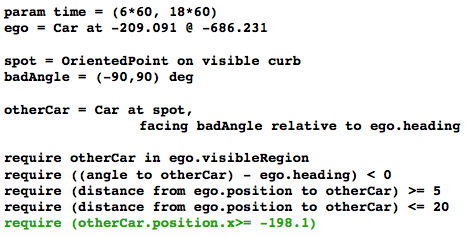}}\quad
  \subfigure[Refined Failure Scenario 1 \scenic program]{\includegraphics[scale=0.5]{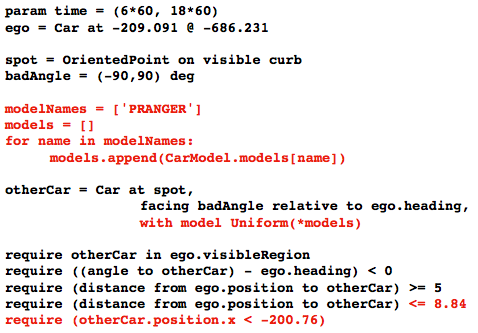}}
  \caption{\label{fig:supp_refined_scenario1}% 
  Refined Success and Failure \scenic programs for Scenario 1, with red parts representing failure inducing rules and green parts representing the success inducing rules as shown in Table \ref{table:correct_rules} and \ref{table:incorrect_rules}}
\end{figure*}

\begin{figure*}
  \centering
  \subfigure[Refined Success Scenario 2 \scenic program]{\includegraphics[scale=0.5]{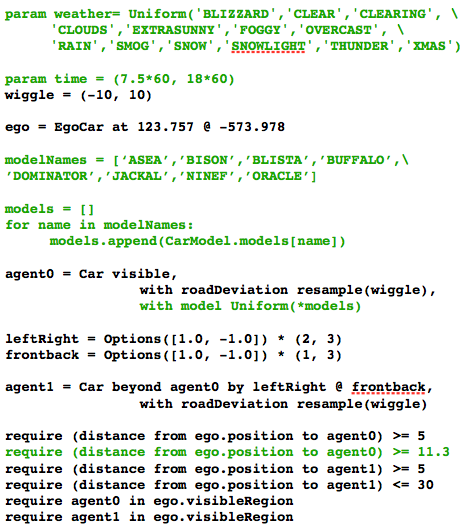}}\quad
  \subfigure[Refined Failure Scenario 2 \scenic program]{\includegraphics[scale=0.5]{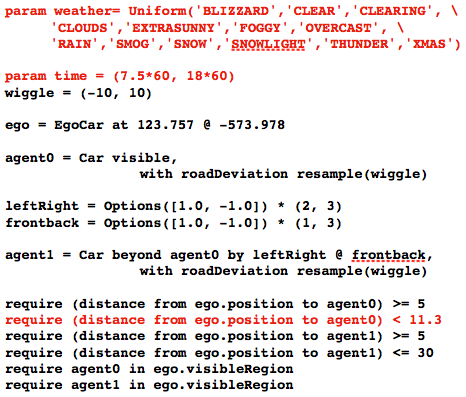}}
  
  \caption{\label{fig:supp_refined_scenario2}%
  Refined Success and Failure \scenic programs for Scenario 2, with red parts representing failure inducing rules and green parts representing the success inducing rules as shown in Table \ref{table:correct_rules} and \ref{table:incorrect_rules}}
\end{figure*}

\begin{figure*}
  \centering
  \subfigure[Refined Success Scenario 3 \scenic program]{\includegraphics[scale=0.4]{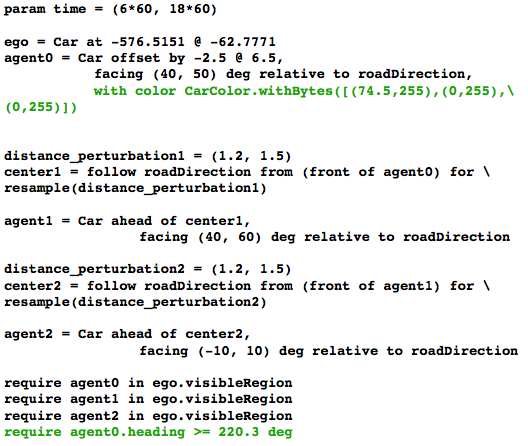}}\quad
  \subfigure[Refined Failure Scenario 3 \scenic program]{\includegraphics[scale=0.4]{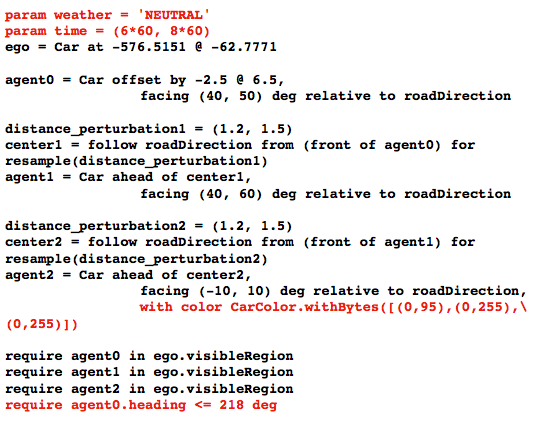}}
  
  \caption{\label{fig:supp_refined_scenario3}%
  Refined Success and Failure \scenic programs for Scenario 3, with red parts representing failure inducing rules and green parts representing the success inducing rules as shown in Table \ref{table:correct_rules} and \ref{table:incorrect_rules}}
\end{figure*}

\begin{figure*}
  \centering
  \subfigure[Refined Success Scenario 4 \scenic program]{\includegraphics[scale=0.45]{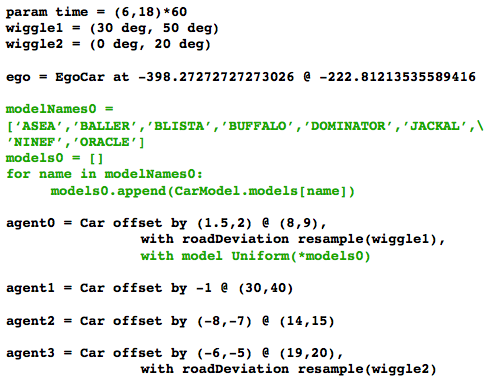}}\quad
  \subfigure[Refined Failure Scenario 4 \scenic program]{\includegraphics[scale=0.45]{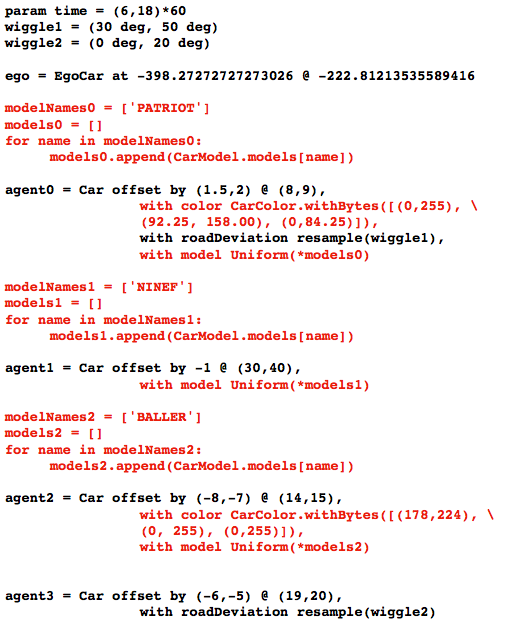}}
  
  \caption{\label{fig:supp_refined_scenario4}%
  Refined Success and Failure \scenic programs for Scenario 4, with red parts representing failure inducing rules and green parts representing the success inducing rules as shown in Table \ref{table:correct_rules} and \ref{table:incorrect_rules}}
\end{figure*}

\begin{figure*}
  \centering
    \includegraphics[width=\textwidth]{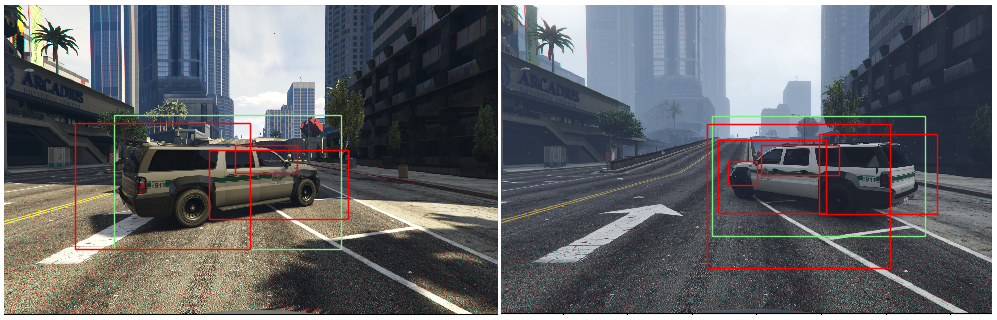}
  \caption{Failure images generated with Failure \scenic Scenario 1 as shown in Figure \ref{fig:supp_refined_scenario1} (b), with ground truth bounding boxes marked in green and prediction bounding boxes in red}
  \label{fig:supp_scenario1_failure_images}
\end{figure*}

\begin{figure*}
  \centering
    \includegraphics[width=\textwidth]{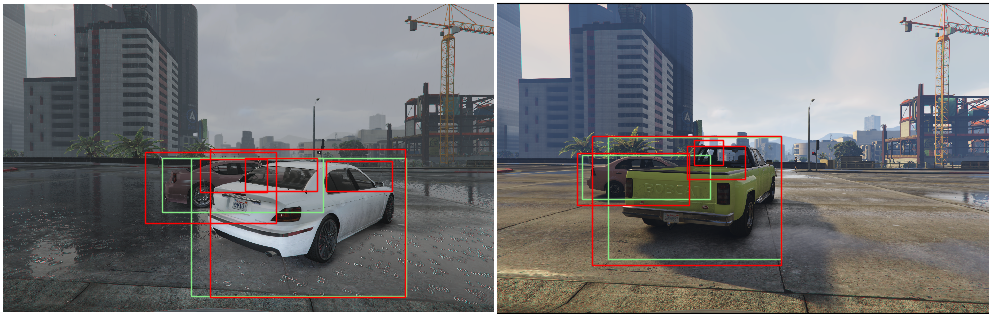}
  \caption{Failure images generated with Failure \scenic Scenario 2 as shown in Figure \ref{fig:supp_refined_scenario2} (b), with ground truth bounding boxes marked in green and prediction bounding boxes in red}
  \label{fig:supp_scenario2_failure_images}
\end{figure*}

\begin{figure*}
  \centering
    \includegraphics[width=\textwidth]{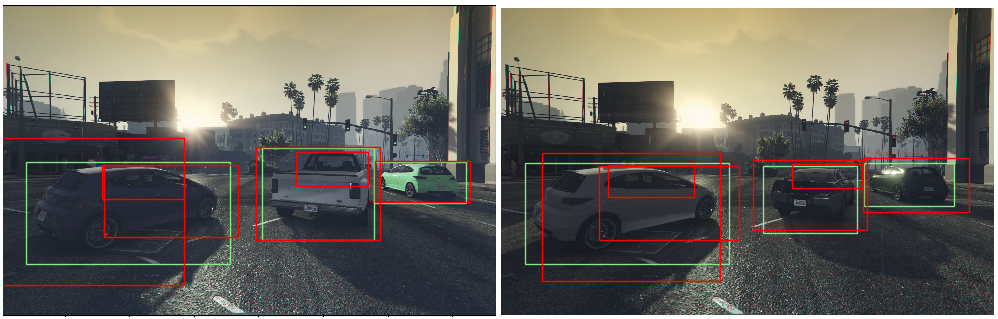}
  \caption{Failure images generated with Failure \scenic Scenario 3 as shown in Figure \ref{fig:supp_refined_scenario3} (b), with ground truth bounding boxes marked in green and prediction bounding boxes in red}
  \label{fig:supp_scenario3_failure_images}
\end{figure*}

\begin{figure*}[t]
  \centering
    \includegraphics[width=\textwidth]{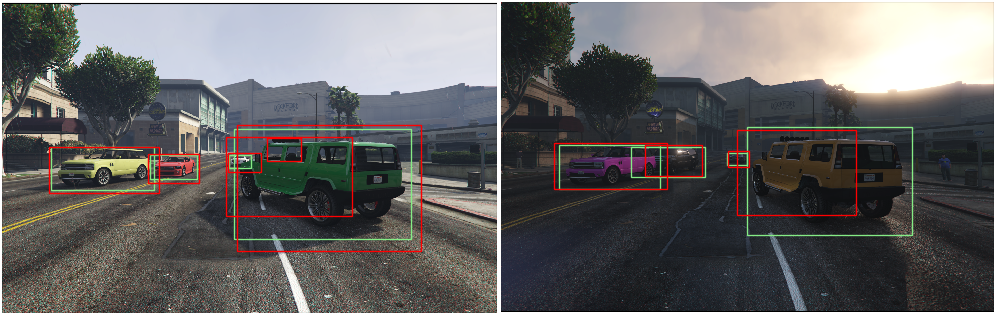}
  \caption{Failure images generated with Failure \scenic Scenario 4 as shown in Figure \ref{fig:supp_refined_scenario4} (b), with ground truth bounding boxes marked in green and prediction bounding boxes in red}
  \label{fig:supp_scenario4_failure_images}
\end{figure*}
\end{appendices}

\end{document}